\documentclass[11pt,a4paper]{article}
\usepackage[hyperref]{emnlp-ijcnlp-2019}
\usepackage{times}
\usepackage{latexsym}
\usepackage{hyperref}
\usepackage{url}
\usepackage{graphicx}
\usepackage{amsmath}
\usepackage{multirow}
\usepackage{booktabs}
\usepackage{amssymb}
\usepackage{csquotes}
\usepackage{blindtext}
\usepackage{enumitem}
\definecolor{blue}{RGB}{0, 93, 170}
\definecolor{green}{RGB}{34,139,34}
\definecolor{red}{RGB}{220,20,60}
\aclfinalcopy 

\title{Deep Reinforcement Learning with Distributional Semantic Rewards \\ for Abstractive Summarization }

\author{Siyao Li$^{1}$\thanks{\enspace Equal contributions.}, Deren Lei$^{1}$\footnotemark[1], Pengda Qin$^2$, William Yang Wang$^1$\\
$^1$University of California, Santa Barbara\\
$^2$Beijing University of Posts and Telecommunications \\
\{siyaoli, derenlei\}@ucsb.edu, qinpengda@bupt.edu.cn, william@cs.ucsb.edu\\
}

\date{}

\begin{document}
\maketitle
\begin{abstract}
Deep reinforcement learning (RL) has been a commonly-used strategy for the abstractive summarization task to address both the exposure bias and non-differentiable task issues. However, the conventional reward \textsc{Rouge-L} simply looks for exact n-grams matches between candidates and annotated references, which inevitably makes the generated sentences repetitive and incoherent. In this paper, instead of \textsc{Rouge-L}, we explore the practicability of utilizing the distributional semantics to measure the matching degrees. With distributional semantics, sentence-level evaluation can be obtained, and semantically-correct phrases can also be generated without being limited to the surface form of the reference sentences. Human judgments on Gigaword and CNN/Daily Mail datasets show that our proposed distributional semantics reward (DSR) has distinct superiority in capturing the lexical and compositional diversity of natural language. 
\end{abstract}
\section{Introduction}

Abstractive summarization is a task of paraphrasing a long article with fewer words. Unlike extractive summarization, abstractive summaries can include tokens out of the article's vocabulary. There exists several encoder-decoder approaches such as attention-based architecture \cite{bahdanau2014neural,rush2015neural,nallapati2016abstractive, chopra2016abstractive}, incorporating graph techniques \cite{moawad2012semantic, ganesan2010opinosis}, and adding pointer generator \cite{vinyals2015pointer,nallapati2016abstractive,see2017get, bello2016neural7}.
However, long sentence generation suffers from exposure bias \cite{bahdanau2016actor} as the error accumulates during the decoding process.

Many innovative deep RL methods \cite{ranzato2015sequence,wu2016google,paulus2017deep,lamb2016professor} are developed to alleviate this issue by providing sentence-level feedback after generating a complete sentence, in addition to optimal transport usage \cite{napoles2012annotated}. However, commonly used automatic evaluation metrics for generating sentence-level rewards count exact n-grams matches and are not robust to different words that share similar meanings since the semantic level reward is deficient.

Currently, many studies on contextualized word representations \cite{peters2018deep, devlin2019bert} prove that they have a powerful capacity of reflecting distributional semantic. In this paper, we propose to use the distributional semantic reward to boost the RL-based abstractive summarization system. Moreover, we design several novel objective functions.

Experiment results show that they outperform the conventional objectives while increasing the sentence fluency. Our main contributions are three-fold: \\
$\bullet$ We are the first to introduce DSR to abstractive summarization and achieve better results than conventional rewards. \\
$\bullet$ Unlike \textsc{Rouge}, our DSR does not rely on cross-entropy loss (XENT) to produce readable phrases. Thus, no exposure bias is introduced. \\
$\bullet$ DSR improves generated tokens' diversity and fluency while avoiding unnecessary repetitions.

\section{Methodology}

\paragraph{Background} While sequence models are usually trained using XENT, they are typically evaluated at test time using discrete NLP metrics such as \textsc{Bleu} \cite{papineni2002bleu}, \textsc{Rouge}\ \cite{lin2004rouge}, \textsc{Meteor} \cite{banerjee2005meteor}. Therefore, they suffer from both the exposure bias and non-differentiable task metric issues. To solve these problems, many resorts to deep RL with sequence to sequence model \cite{paulus2017deep,ranzato2015sequence,ryang2012framework}, where the learning agent interacts with a given environment. However, RL models have poor sample efficiency and lead to very slow convergence rate. Therefore, RL methods usually start from a pretrained policy, which is established by optimizing XENT at each word generation step.
\begin{equation}
    L_{\text{XENT}} = -\sum_{t=1}^{n'}\log P(y_t|y_1, \dots, y_{t-1}, x).
\end{equation}
Then, during RL stage, the conventional way is to adopt self-critical strategy to fine-tune based on the target evaluation metric,
\begin{align}
    \label{Lrouge}
    L_{\text{RL}} = &
            \sum_{t=1}^{n'}\log P(\hat{y_t}|\hat{y_1}, \dots, \hat{y_{t-1}}, x) \\
    &\times (r_{metric}(y^b)-r_{metric}(\hat{y})) \nonumber
\end{align}

\paragraph{Distributional Semantic Reward} During evaluating the quality of the generated sentences, \textsc{Rouge} looks for exact matches between references and generations, which naturally overlooks the expression diversity of the natural language. In other words, it fails to capture the semantic relation between similar words. To solve this problem, distributional semantic representations are a practical way. Recent works on contextualized word representations, including ELMO \citep{peters2018deep}, GPT \citep{radford2018improving}, BERT \citep{devlin2019bert}, prove that distributional semantics can be captured effectively. Based on that, a recent study, called \textsc{BERTScore} \citep{zhang2019bertscore}, focuses on sentence-level generation evaluation by using pre-trained BERT contextualized embeddings to compute the similarity between two sentences as a weighted aggregation of cosine similarities between their tokens. It has a higher correlation with human evaluation on text generation tasks comparing to existing evaluation metrics. 

In this paper, we introduce it as a DSR for deep RL.
The \textsc{BERTScore} is defined as:
\begin{equation}
    R_{\text{BERT}} = \frac{ \sum_{y_i \in y} \text{idf}(y_i) \max_{\hat{y}_j \in \hat{y}} \mathbf{y_i}^\top \mathbf{\hat{y}_j}} {\sum_{y_i \in y} \text{idf}(y_i)}
\end{equation}
\begin{equation}
    P_{\text{BERT}} = \frac{ \sum_{y_j \in y} \text{idf}(y_j) \max_{\hat{y}_i \in \hat{y}} \mathbf{y_j}^\top \mathbf{\hat{y}_i}} {\sum_{y_j \in y} \text{idf}(y_j)} \\
\end{equation}
\begin{equation}
    F_{\text{BERT}} = 2\frac{R_{\text{BERT}} \cdot P_{\text{BERT}}}{R_{\text{BERT}}+P_{\text{BERT}}}
\end{equation}
where $\mathbf{y}$ and $\mathbf{\hat{y}}$ represent BERT contextual embeddings of reference word y and candidate word $\hat{y}$, respectively. The function idf($\cdot$) calculates inverse document frequency (idf). In our DSR, we do not use the idf since \citeauthor{zhang2019bertscore} \shortcite{zhang2019bertscore} requires to use the entire dataset including test set for calculation. Besides, \textsc{Rouge} do not use similar weight, so we do not include idf for consistency.

\section{Experimental Setup}

\subsection{Datasets}
\paragraph{Gigaword corpus} It is an English sentence summarization dataset based on annotated Gigaword \cite{napoles2012annotated}. A single sentence summarization is paired with a short article. We use the OpenNMT provided version It contains 3.8M training, 189k development instances. We randomly sample 15k instances as our test data.

\paragraph{CNN/Daily Mail dataset} It consists of online news articles and their corresponding multi-sentence abstracts \cite{hermann2015teaching, nallapati2016abstractive}. We use the non-anonymized version provided by \citet{see2017get}, which contains 287k training, 13k validation, and 11k testing examples. We truncate the articles to 400 tokens and limit the summary lengths to 100 tokens. 

\subsection{Pretrain}
We first pretrain a sequence-to-sequence model with attention using XENT and then select the best parameters to initialize models for RL. Our models have 256-dimensional hidden states and 128-dimensional word embeddings and also incorporates the pointer mechanism \cite{see2017get} for handling out of vocabulary words. 

\subsection{Baseline}
In abstractive summarization, \textsc{Rouge} \cite{lin2004rouge} is a common evaluation metric to provide a sentence-level reward for RL. However, using \textsc{Rouge} as a pure RL objective may cause too many repetitions and reduced fluency in outputs. \citet{paulus2017deep} propose a hybrid learning objective that combines XENT and  self-critical \textsc{Rouge} reward \cite{paulus2017deep}.
\begin{equation}
    \label{eq:baseline}
    L_{\text{baseline}} = \gamma L_{\text{rouge}} + (1-\gamma)L_{\text{XENT}}
\end{equation}
where $\gamma$ is a scaling factor and the F score of \textsc{Rouge-L} is used as the reward to calculate $L_{\text{Rouge}}$. In our experiment, we select $\gamma = 0.998$ for Gigaword Corpus and $\gamma = 0.9984$ for CNN/Daily Mail dataset.
\footnote{$\gamma$ is very close to 1 due to the difference in the scales of the two loss functions. For Gigaword Corpus, we tune the $\gamma$ on the development set. For CNN/Daily Mail, we use the same $\gamma$ as \citet{paulus2017deep} does.}
Note that we do not avoid repetition during the test time as \citet{paulus2017deep} do, because we want to examine the repetition of sentence directly produced after training.

\subsection{Proposed Objective Functions}
Inspired by the above objective function \cite{paulus2017deep}, we optimize RL models with a similar loss function as equation~\ref{Lrouge}. Instead of \textsc{Rouge-L}, we incorporate \textsc{BERTScore}, a DSR to provide sentence-level feedback. 

In our experiment, $L_{\text{DSR}}$ is the self-critical RL loss (equation~\ref{Lrouge}) with $F_{\text{BERT}}$ as the reward. We introduce the following objective functions: 

\paragraph{DSR+\textsc{Rouge}:} A combined reward function of \textsc{Rouge} and $F_{\text{BERT}}$
\begin{equation}
    \label{l1}
    L_{1} = \gamma L_{\text{DSR}} + (1-\gamma)L_{\text{rouge}}
\end{equation}
In our experiment, we select $\gamma = 0.5$ for both datasets to balance the influence of two reward functions.
\paragraph{DSR+XENT:} $F_{\text{BERT}}$ reward with XENT to make the generated phrases more readable.
\begin{equation}
    \label{l2}
    L_{2} = \gamma' L_{\text{DSR}} + (1-\gamma')L_{\text{XENT}}
\end{equation}
In our experiment, we select $\gamma' = 0.998$ for Gigaword Corpus and $\gamma' = 0.9984$ for CNN/Daily Mail dataset.
\paragraph{DSR:} Pure $F_{\text{BERT}}$ objective function without any teacher forcing.
\begin{equation}
    \label{l3}
    L_{3} = L_{\text{DSR}}
\end{equation}

\begin{table*}[h!]
    \centering
    \small
    \begin{tabular}{l|cccc|cccc}
    \toprule
        & \multicolumn{4}{c}{\textbf{Gigawords}} & \multicolumn{4}{c}{\textbf{CNN/Daily Mail}} \\
       \cmidrule{2-9}
       \textbf{Model} & $F_{\text{BERT}}$  & $P_{\text{BERT}}$ & $R_{\text{BERT}}$ & \textsc{Rouge} & $F_{\text{BERT}}$  & $P_{\text{BERT}}$ & $R_{\text{BERT}}$ & \textsc{Rouge}\\ 
        \midrule
        XENT & 65.78 & 67.61 & 64.53 & 40.77 & 62.77 & 62.18 & 63.79 & 29.46\\
        \textsc{Rouge} & 61.46 & 60.89 & 62.70 & 42.73 & 60.11 & 59.31 & 61.33 & \textbf{33.89}\\
        \textsc{Rouge}+XENT & 66.50 & 67.24 & 66.28 & 42.72 & 61.38 & 61.07 & 62.17 & 33.07\\
        \midrule
        \textsc{DSR}+\textsc{Rouge} & 66.48 & 66.79 & 66.65 & \textbf{42.95} & 65.01 & 65.92 & 64.56 & 33.61\\
        \textsc{DSR}+XENT  & 67.02 & \textbf{67.69} & 66.85 & 42.29 & 66.64 & 66.06 & 67.63 & 31.28\\
        \textsc{DSR}  & \textbf{67.06} & 67.34 & \textbf{67.28} & 41.73 & \textbf{66.93} & \textbf{66.27} & \textbf{67.98} & 30.96\\
        \bottomrule
    \end{tabular}
    \caption{Results on Gigawords and CNN/Daily Mail for abstractive summarization.  Upper rows show the results of baselines. Rouge stands for the F score of Rouge-L.}
    \label{tab:onto1}
\end{table*}

\section{Results}
For the abstractive summarization task, we test our models with different objectives on the Gigaword and CNN/Daily Mail datasets. We choose $F_{\text{BERT}}$ and \textsc{Rouge-L} as our automatic evaluation metrics. For multi-sentence summaries in CNN/Daily Mail dataset, we calculate \textsc{Rouge-L} and $F_{\text{BERT}}$ after concatenating the original abstract sentences or their BERT contextualized embeddings. The results are shown in Table~\ref{tab:onto1}. The model DSR+\textsc{Rouge} (equation~\ref{l1}) achieves the best \textsc{Rouge} performance on Gigaword and similar performance comparing to \textsc{Rouge} model on CNN/Daily Mail. Training with DSR (equation~\ref{l3}) and DSR+XENT (equation~\ref{l2}) can obtain the best BERTScores as expected. It is also expected that \textsc{Rouge} model will obtain worse BERTScores as simply optimizing \textsc{Rouge} will generate less readable sentences \cite{paulus2017deep}; however, DSR model without XENT as a teacher forcing can improve the performance of pretrained model in both $F_{\text{BERT}}$ and \textsc{Rouge-L} scale.

Note that DSR model's \textsc{Rouge-L} is high in training time but does not have a good generalization on test set, and \textsc{Rouge-L} is not our target evaluation metrics.
In the next section, we will do human evaluation to analyze the summarization performance of different reward systems.

\begin{table*}
    \centering
    \small
    \begin{tabular}{c|c|ccc|ccc}
    \toprule
        \multirow{2}{*}{\textbf{Task}} &  \multirow{2}{*}{\textbf{Model v. base}} & \multicolumn{3}{c}{\textbf{Gigawords}} & \multicolumn{3}{c}{\textbf{CNN/Daily Mail}} \\
                            &    &  Win & Lose & Tie &  Win & Lose & Tie\\ 
        \midrule
        \multirow{2}{*}{Relevance} & \textsc{DSR} + XENT & 31.0\% & 25.2\% & 43.8\% & 51.1\% & 34.1\% & 14.8\%\\
                                   & \textsc{DSR} & 45.4\% & 27.2\% & 27.4\% & 48.7\% & 38.5\% & 12.8\%\\
                                     \midrule
        \multirow{2}{*}{Fluency} & \textsc{DSR} + XENT & 40.2\% & 19.6\% & 40.2\% & 55.8\% & 28.5\% & 15.6\%\\
                                 & \textsc{DSR} & 45.4\% & 28.8\% & 25.8\% & 54.0\% & 31.7\% & 14.3\%\\
        \bottomrule
    \end{tabular}
    \caption{Human evaluation results on Gigaword and CNN/Daily Mail for abstractive summarization. \textsc{Rouge}+XENT is the selected baseline model.}
    \label{tab:onto2}
\end{table*}

\section{Human Evaluation}
\begin{figure}[t]
\centering
\includegraphics[width=7.5cm]{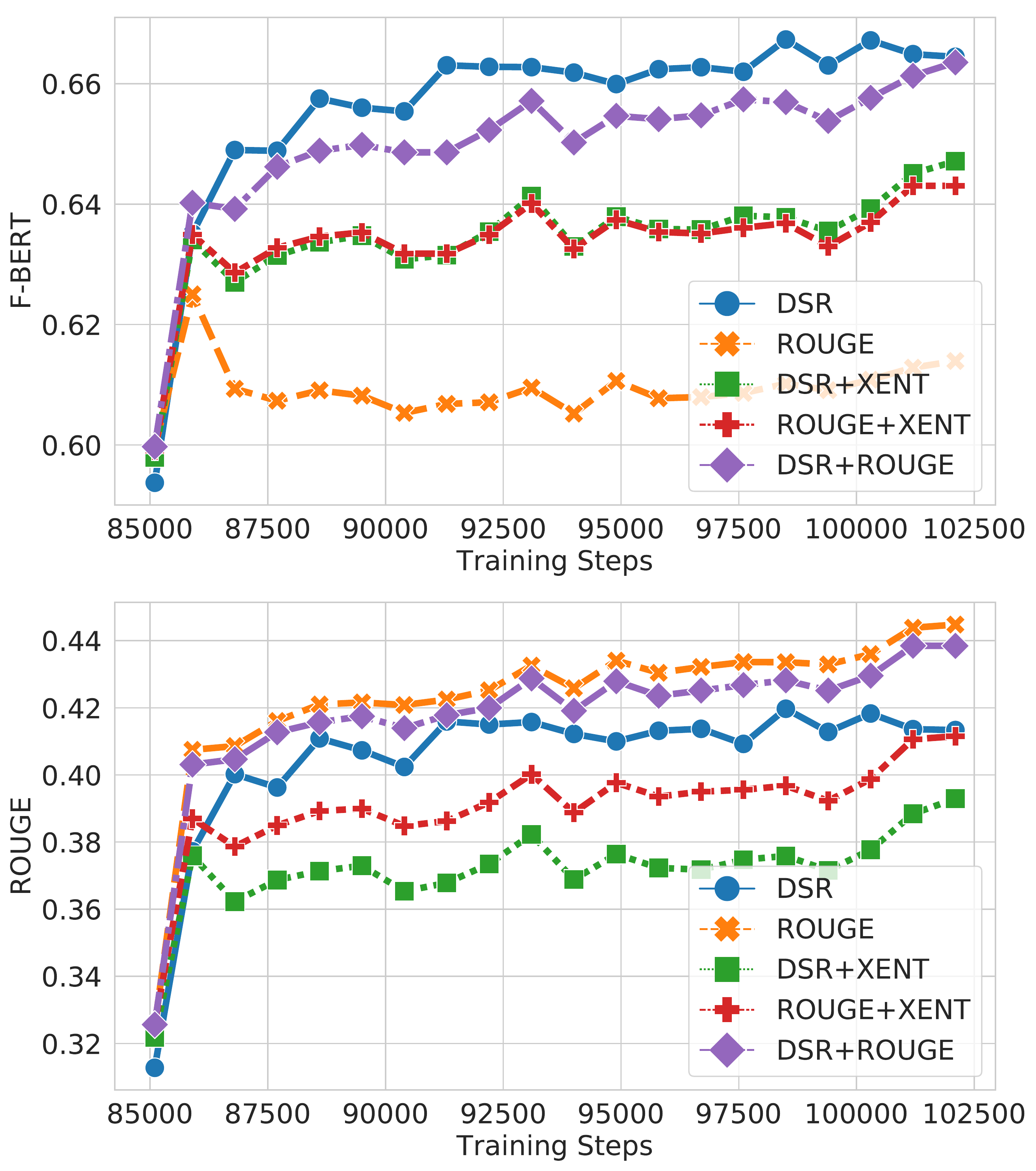}
\caption{$F_{\text{BERT}}$ (left) and ROUGE (right) on Gigaword Corpus during training time after pretraining with XENT. Note that ROUGE is not our target evaluation metrics.}
\label{fig:score}
\end{figure}
We perform human evaluation on the Amazon Mechanical Turk to assure the benefits of DSR on output sentences' coherence and fluency. We randomly sample 500 items as an evaluation set using uniform distribution. During the evaluation, given the context without knowing the ground truth summary, judges are asked to decide two things between generations of \textsc{Rouge}+XENT and DSR+XENT: 1. which summary is more relevant to the article. 2. which summary is more fluent. Ties are permitted. A similar test is also done between \textsc{Rouge}+XENT and DSR model.
As shown in Table~\ref{tab:onto2}, DSR and DSR+XENT models improve the relevance and fluency of generated summary significantly. In addition, using pure DSR achieve better performances on Gigaword Corpus and comparable results as DSR+XENT on CNN/Daily Mail. While \citeauthor{paulus2017deep}'s \shortcite{paulus2017deep} objective function requires XENT to make generated sentence readable, our proposed DSR does not require XENT and can limit the exposure bias originated from it.

\section{Analysis}
\paragraph{Diversity}
Other than extractive summarization, abstractive summarization allows more degrees of freedom in the choice of words. While simply selecting words from the article made the task easier to train, higher action space can provide more paths to potentially better results \cite{Nema_2017}. Using the DSR as deep RL reward will support models to choose actions that are not n-grams of the articles. In Table~\ref{tab:big}, we list a few generated samples on Gigaword Corpus. In our first example in Table~\ref{tab:big}, the word ``sino-german" provides an interesting and efficient way to express the relation between China and Germany. $F_{\text{BERT}}$ is also improved by making this change. In addition, the second example in Table~\ref{tab:big} shows that RL model with DSR corrects the sentence' grammar and significantly improves the $F_{\text{BERT}}$ score by switching ``down" to an unseen word ``drops". On the other hand, when optimizing DSR to improve the diversity of generation, some semantically similar words may also be generated and harm the summarization quality as shown in the third example in Table~\ref{tab:big}. The new token ``wins" reduces the scores of both metrics. We also evaluate the diversity of a model quantitively by averaging the percentage of out-of-article n-grams in generated sentences. Results can be found in Table~\ref{tab:analysis}. The DSR model achieves the highest diversity.
\begin{table}
    \centering
    \small
    \begin{tabular}{c|c|c}
    \toprule
        \multirow{2}{*}{\textbf{Model}}  & \multicolumn{1}{c}{\textbf{Gigawords}} & \multicolumn{1}{c}{\textbf{CNN/Daily Mail}} \\
           & Rep(\%) Div(\%) & Rep(\%) Div(\%)\\ 
        \midrule
        \textsc{Rouge}+XENT & 11.57 \quad 16.33 & 66.82 \quad 21.32 \\
        \midrule
        \textsc{DSR}+\textsc{Rouge} & 18.42 \quad 15.97 & 50.41 \quad 34.70 \\
        \textsc{DSR}+XENT  & 20.15 \quad 16.24 & 25.98 \quad 31.80 \\
        \textsc{DSR}  & \textbf{  7.20} \quad \textbf{ 19.17} & \textbf{22.03} \quad \textbf{41.35} \\
        \bottomrule
    \end{tabular}
    \caption{Qualitative analysis on repetition(Rep) / diversity(Div). They are calculated by the percentage of repeat/out-of-article n-grams (unigrams for Gigaword and 5-grams for CNN/Daily Mail) in generated sentences. }
    \label{tab:analysis}
\end{table}
\begin{table*}[]
    \centering
    \small
    \begin{tabular}{c|l}
    \toprule 
        1) Context & \multirow{3}{13cm}{economic links between china and germany will be at the center of talks here next month between li peng and chancellor helmut kohl when the chinese premier makes a week-long visit to germany , sources on both sides said .} \\
        & \\ 
        & \\
        Groundtruth & li peng 's visit to germany to focus on economic ties \\
        \textsc{Rouge}+XENT & chinese premier to visit germany next month \textcolor{green}{\{ $F_{\text{BERT}}$: 49.48, \textsc{Rouge}:\textbf{18.64}\}}\\
        \textsc{DSR}+XENT & chinese premier to visit germany next month \textcolor{green}{\{ $F_{\text{BERT}}$: 49.48, \textsc{Rouge}:\textbf{18.64}\}} \\
        \textsc{DSR} & \textbf{\textcolor{blue}{sino-german}} economic \textbf{\textcolor{blue}{links}} to be at center of talks \textbf{\textcolor{blue}{with}} germany \textcolor{green}{\{$F_{\text{BERT}}$: \textbf{50.77}, \textsc{Rouge}:17.33\}} \\ \midrule
        
        2) Context & \multirow{3}{13cm}{the sensitive index dropped by \#.\#\# percent friday on the bombay stock exchange -lrb- bse -rrb- following brisk inquiries from operators that generally pressed sales to square up positions on the last day of the current settlement .} \\
        & \\
        & \\
        Groundtruth & sensex falls in bombay stock exchange \\
        \textsc{Rouge}+XENT & sensitive index down on bombay stock exchange \textcolor{green}{\{$F_{\text{BERT}}$: 70.03, \textsc{Rouge}:\textbf{45.62}\}} \\
        \textsc{DSR}+XENT & sensitive index \textbf{\textcolor{blue}{drops}} on bombay stock exchange \textcolor{green}{\{$F_{\text{BERT}}$: \textbf{73.92}, \textsc{Rouge}:\textbf{45.62}\}} \\
        \textsc{DSR} & sensitive index \textbf{\textcolor{blue}{drops}} on bombay stock exchange \textcolor{green}{\{$F_{\text{BERT}}$: \textbf{73.92}, \textsc{Rouge}:\textbf{45.62}\}} \\ \midrule
        
        3) Context & \multirow{2}{13cm}{belgium 's rik verbrugghe took victory in tuesday 's prologue stage of the tour de romandie , with a confident ride through the streets of geneva 's old town .} \\
        & \\
        Groundtruth & verbrugghe takes prologue victory in tour de romandie \\
        \textsc{Rouge}+XENT &  verbrugghe \textbf{\textcolor{blue}{takes}} victory in prologue stage of tour de romandie \textcolor{green}{\{$F_{\text{BERT}}$: 91.41, \textsc{Rouge}:75.93\}}\\
        \textsc{DSR}+XENT & verbrugghe \textbf{\textcolor{blue}{takes}} victory in tour de romandie prologue stage \textcolor{green}{\{$F_{\text{BERT}}$: \textbf{91.56}, \textsc{Rouge}:\textbf{81.79}\}} \\
        \textsc{DSR} & verbrugghe \textbf{\textcolor{blue}{wins}} victory in tour de romandie prologue stage \textcolor{green}{\{$F_{\text{BERT}}$: 89.45, \textsc{Rouge}:70.10\}} \\ \midrule
        
        4) Context & \multirow{2}{13cm}{european finance ministers on saturday urged swedes to vote `` yes '' to adopting the euro , saying entry into the currency bloc would be good for sweden and the rest of europe .} \\
        & \\
        Groundtruth & european finance ministers urge swedes to vote yes to euro \\
        \textsc{Rouge}+XENT & \textbf{\textcolor{blue}{eu}} finance ministers urge swedes to vote yes to euro \textcolor{green}{\{$F_{\text{BERT}}$: \textbf{96.63}, \textsc{Rouge}:\textbf{93.47}\}} \\
        \textsc{DSR}+XENT & \textbf{\textcolor{blue}{eu}} finance ministers urge swedes to vote yes to adopting euro \textcolor{green}{\{$F_{\text{BERT}}$: 90.38, \textsc{Rouge}:88.89\}} \\
        \textsc{DSR} & \textbf{\textcolor{blue}{eu}} finance ministers urge swedes to vote yes to \textcolor{red}{euro euro} \textcolor{green}{\{$F_{\text{BERT}}$: 95.05, \textsc{Rouge}:\textbf{93.47}\}} \\ \bottomrule
        
    \end{tabular}
    \caption{Qualitative analysis of generated samples on Gigaword corpus. Generated words that do not appear in the context are marked blue. Repeated words are marked red. The first two examples represent DSR's generated tokens are more diverse. However, it may suffer from problems as shown in example 3 and 4.}
    \label{tab:big}
\end{table*}

\paragraph{Repetition}
Repetition will lead to lower $F_{\text{BERT}}$ as shown in the last example in Table~\ref{tab:big}. Using DSR reduces the probability of producing repetitions. The average percentage of repeated n-grams in generated sentences are presented in the Table~\ref{tab:analysis}. As shown in this table, unlike \textsc{Rouge}, the DSR model can achieve high fluency without XENT; moreover, it produces the fewest repetitions among all the rewards. Table~\ref{tab:big} gives an example that DSR produces a repeated word (from example 4), but it does not reflect the overall distribution of repeated word generation for all evaluated models.

\section{Conclusion}
This paper demonstrates the effectiveness of applying the distributional semantic reward to reinforcement learning in abstractive summarization, and specifically, we choose \textsc{BERTScore}. Our experimental results demonstrate that we achieve better performance on Gigaword and CNN/Daily Mail datasets. Besides, the generated sentences have fewer repetitions, and the fluency is also improved. Our finding is aligned to a contemporaneous study \cite{wieting2019beyond} on leveraging semantic similarity for machine translation.

\bibliography{citation}
\bibliographystyle{acl_natbib}

\end{document}